%% file: Card-JAMIA2012.tex
\documentclass{amia}
\usepackage{graphicx}
\usepackage{paralist}
\usepackage{booktabs}

\usepackage[numbers,super]{natbib}

\begin{document}

\title{Supervised Laplacian Eigenmaps with Applications in
Clinical Diagnostics for Pediatric Cardiology}

\author{Thomas Perry$^1$, ~Hongyuan Zha$^1$, Patricio Frias$^{2,3}$, Dadan Zeng$^{4}$, Mark Braunstein$^{1}$ }

\institutes{
    $^1$ Georgia Institute of Technology, Atlanta, GA \\
    $^2$ Children's Healthcare of Atlanta Sibley Heart Center, Atlanta, GA \\
    $^3$ Emory University School of Medicine, Atlanta, GA\\
    $^4$ East China Normal University, Shanghai, China\\
}

\maketitle

\input{introduction.tex}
\input{background.tex}
\input{methods.tex} 
\input{results.tex}
\input{discussion.tex}

\bibliographystyle{ieeetr}
\bibliography{bibliography}

\end{document}

%% file: introduction.tex
\section*{Abstract}
\textit{
Electronic health records contain rich textual data which possess critical predictive information for machine-learning based diagnostic aids.  However many traditional machine learning methods fail to simultaneously integrate both vector space data and text.  We present a supervised method using Laplacian eigenmaps to augment existing machine-learning methods with low-dimensional representations of textual predictors which preserve the local similarities.  The proposed implementation performs alternating optimization using gradient descent.  For the evaluation we applied our method to over 2,000 patient records from a large single-center pediatric cardiology practice to predict if patients were diagnosed with cardiac disease.  Our method was compared with latent semantic indexing, latent Dirichlet allocation, and local Fisher discriminant analysis.  The results were assessed using AUC, MCC, specificity, and sensitivity.  Results indicate supervised Laplacian eigenmaps was the highest performing method in our study, achieving 0.782 and 0.374 for AUC and MCC respectively.  SLE showed an increase in 8.16\% in AUC and 20.6\% in MCC over the baseline which excluded textual data and a 2.69\% and 5.35\% increase in AUC and MCC respectively over unsupervised Laplacian eigenmaps.  This method allows many existing machine learning predictors to effectively and efficiently utilize the potential of textual predictors.
}

\section*{Introduction}
Appropriate resource utilization is a key element with respect to successful reform of the US healthcare delivery system\cite{schreiber95,memtsoudis12,bodenheimer99}. Electronic health records in combination with machine learning techniques are being employed to develop diagnostic aids to improve patient outcomes and operational efficiency\cite{kukar99,ahmed09,natt05}.  The efficacy of these methods is often limited by data quality\cite{cortes95}, quantity, and structure.  Despite this, diagnostic aids commonly employ machine learning methods\cite{cortessvm95,tib94,schol98} which rely on vector space representations  of data and exclude unstructured data such as text, which may contain critical predictive information.  Clinical pathways often produce concise textual notes from patients and physicians.  For example, they may include a patient's description of his/her symptoms or the outcomes of diagnostic exams.  Just as this information is critical for physicians to assess pathology and recommend appropriate intervention, incorporating textual data may be vital to developing effective machine-learning based diagnostic aids\cite{rosbe98,bhatia10,turchin08}.  Previous research on developing diagnostic aids for pediatric cardiology has excluded textual data\cite{perryAmia}.  We present a novel method to augment existing machine learning predictors by embedding textual predictors into a low-dimensional Euclidean space which preserves the local similarities among textual predictors.  Our contribution simultaneously learns the parameters of the base predictor and uses Laplacian eigenmaps\cite{belkin02} to optimize the embedding of concise documents. By performing joint optimization, the low-dimensional representation of textual data is constructed to maximize the model's generalized predictive performance.  This method allows existing machine learning methods to extract the previously unutilized potential of textual data in an effective and efficient manner.

This method was evaluated on data from a pediatric cardiology network containing over 2,000 patient records which were diagnosed with major, minor, or no cardiac disease.  We combined the proposed method with logistic regression as the base classifier.  Results show improved performance over alternative methods of dimensionality reduction for textual data such as latent semantic indexing, latent Dirichlet allocation, and local Fisher discriminant analysis.  By effectively leveraging textual predictors from EHRs, we demonstrate the need for supervised machine learning diagnostic aids which incorporate both numerical and textual data.

%% file: background.tex
\section*{Background}
While current research utilizing dimensionality reduction for unstructured text in clinical settings is limited, much of the work fails to integrate both textual and numerical data into a unified model. Halpren et al.\cite{halpern12} examine triage notes to classify if a patient has an infection or will be admitted to the ICU. Salleb-Aouissi  et al.\cite{sallab} focuses on applying LDA\cite{blei03} to pediatric notes to better analyze infant colic.  Perotte et al.\cite{perotte} present a method to assign ICD-9 codes from discharge summaries by developing a new latent Dirichlet allocation (LDA) model for hierarchical prediction.

Unsupervised text dimensionality reduction methods focus on discovering a latent structure which maximizes the likelihood of the text corpus or minimize an error function, while supervised methods utilize labelled data to construct a low-dimensional representation for a specific task.  For example, the unsupervised low-dimensional representation of text-based movie reviews may primarily focus on features such as genre.  However, a supervised low-dimensional representation of movie reviews with the task of predicting user sentiment may focus on words such as "excellent", "good", or "horrible".  Our method differs from previous work in that it incorporates numerical data in addition to text.

Common unsupervised dimensionality reduction techniques such as singular value decomposition (SVD)\cite{trefethen} methods and LDA exclusively utilize textual data or vector representations of textual data. The SVD computes the best rank-$k$ approximation, as defined by the Frobenius norm.  Various implementations of this method are latent semantic indexing (LSI) and principle component analysis (PCA).  These procedures decrease noise while maintaining the most important semantic information.  LDA is an alternative unsupervised dimensionality technique utilizing topic models.  LDA is a generative process which models the words of a document and attempts to infer the topics to maximize the likelihood of the collection.  The low dimensional representation is specified by the distribution $\theta_d$, the proportion of a document which corresponds to each topic.

Supervised dimensionality methods often produce better low-dimensional representations for specific learning tasks.  Many supervised approaches use topic modelling, which traditionally excludes numerical data.  Topic modelling is typically applied to long text documents whereas our method is designed for concise documents such as clinical notes. However, research by Halpren et al.\cite{halpern12} has shown potential for supervised topic models for dimensionality reduction of concise clinical textual data (10-30 words).  Examples of successful techniques are supervised LDA (sLDA)\cite{blei07} and maximum entropy discrimination LDA (MedLDA)\cite{zhu09}.  sLDA augments LDA by adding a response variable to each document and making the learning task to maximize the joint likelihood of the data and response variables. MedLDA trains supervised topic models to estimate topic representations using the max-margin principle.  An alternative supervised dimensionality reduction method is local Fisher discriminant analysis (LFDA)\cite{sug06,sug07}.  LFDA effectively combines the ideas of Fisher disciminant analysis and locality preserving projection.  LFDA maximizes between-class separability and preserves within-class local structure at the same time.

%% file: methods.tex
\section*{Text-Based Similarity Measure}

The complexities of computing the similarity of textual data arises from its unstructured nature.  Text documents may vary in the number of statements, vocabulary, sentence composition, and include human error.  For example the text entries "CP with activity" and "Exercise induced chest pain" share a similar semantic meaning but do not share a single word.  Advanced techniques must be employed to account for complex textual relationships such as abbreviations, acronyms, misspellings, and synonyms.  We propose a similarity measure designed for very short clinical text documents based on a field matching technique for record linkage \cite{minton2005}.  Our key assumptions are 1) a document is composed of one or more statements (Figure 1), 2) the similarity between documents is determined by the similarity of their statements, and 3) the similarity between statements is determined through a transformation process.
 
Before computing the similarity between documents, textual data must first be normalized to reduce error and variations in similarity values. As we cannot enumerate all possible normalization techniques and the required techniques are defined by the dataset, we do not present a specific set of normalization rules. The normalization methods we utilized includes converting textual data to lower case and removing excess white-space, stop words, and punctuation which was not be used as a delimiter.  We denote the set of normalized documents as $T = [T_1,T_2,\ldots,T_m]$.

\begin{figure}
\centering
\includegraphics[width=16.5cm]{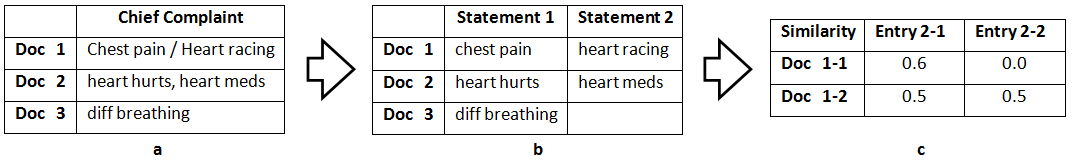}
\caption{Document similarity process. (a) sample short clinical texts. (b) document normalization and segmentation into statements. (c) statement similarities for documents 1 and 2.}
\end{figure}

We assume a document $T_i$ contains one or more statements.  For example an EHR may list "Chest pain / Heart racing" as the patient's chief complaint (Figure 1a).  Each document $T_i$  is split into statements using delimiters such that  $T_i = [T_{i1}, T_{i2},\ldots,T_{ir_i}]$ and consists of $r_i$ statements (Figure 1b). Applying this procedure to our previous example results in the following statements ["chest pain", "heart racing"].  The set of delimiters is defined by the dataset and may require advanced machine learning techniques to correctly segment text values.  Common delimiters are commas, periods, and forward-slashes. Each statement is composed of a set of tokens, either words or numbers.  

In this method the similarity between documents is determined by comparing the similarity of their statements.  We assume statement ordering within documents is irrelevant. Therefore there exists multiplex ways to compare the pairs of statements within two documents (Figure 1c).  However, we assume for this application that the number of statements is small (i.e. $\mid T_i \mid \leq 6$).  This limits the number of comparisons to at most 36.

The key to determining statement similarities is the use of transformations to relate two statements. For example, Figure 2 shows how the statements "CP with activity" and "exercise induced chest pain" are related.  "CP" is an acronym for "chest pain" while "exercise" and "activity" are synonyms. While the word "with" is commonly considered a stop word, it was not removed for readability purposes. The set of  textual transformations, $\kappa$, employed in this method are Equal, Synonym, Misspelling, Abbreviation, Prefix, Acronym, Concatenation, Suffix, and Missing.  This method was taken from work by Minton et al.\cite{minton2005} on field matching.  These transformations are generic and can be applied to other domains outside of medical informatics. However, some transformations may require domain specific customization.  For example, the "Synonym" and "Acronym" transformations may require a list of appropriate terms developed for each domain.

Given two statements, $T_{ij}$ and $T_{kl}$, a transformation $\kappa_u$ (ex. Synonym or Acronym) maps tokens in $T_{ij}$ to tokens in $T_{kl}$.  The transformation "Equal" implies two tokens are identical while the transformation "Missing" maps a token to the empty set.  A transformation graph is built to relate two statements, $T_{ij}$ to $T_{kl}$ where each transformation must be valid.  For example, tokens which are not synonyms may not be related by the Synonym transform.   Every token in $T_{ij}$ and $T_{kl}$ must participate in a transformation for the graph to be considered complete and each token must only participate in a single transformation to be considered consistent.

We represent a transformation graph by a transformation vector $c$, a $\mid \kappa \mid \times 1$ vector of non-negative integers such that $c_u$ corresponds to the number of times transformation $\kappa_u$ was used within the transformation graph.  As there are potentially a large set of possible transformation graphs, we let $\mathcal{C}_{ijkl}$ denote the set of all possible complete consistent transformation vectors for comparing statements $T_{ij}$ and $T_{kl}$.  We assign a transform similarity value $\omega_u$ to each transform $\kappa_u$, where $0 \le \omega_u \le 1$.  Naturally a higher transform similarity value indicates that a transform implies two tokens are more similar.  For example, the similarity score for Equal would naturally be higher than Synonym or Missing and equal to one.  The transformation similarity vector is defined by the user or through an automated process.  Statement similarity is defined as follows  

\begin{figure}
\centering
\includegraphics[width=6.5cm]{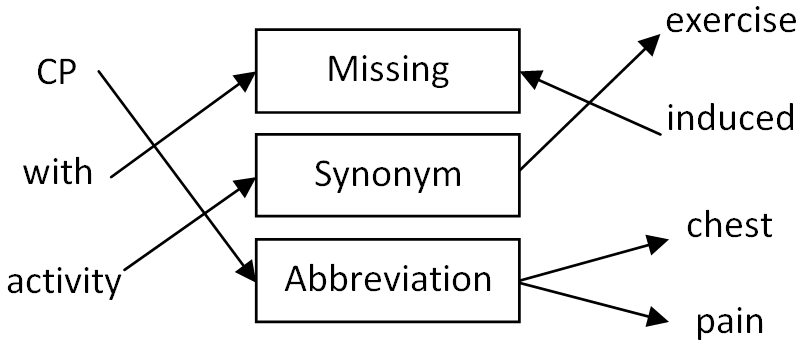}
\caption{Transformation graph illustration}
\end{figure}

\begin{equation}
S_{ijkl} =   \max  ~ \left \{ \frac{\omega ^T c}{\sum c}  \mid c \in \mathcal{C}_{ijkl} \right \}
\end{equation}

When the similarity matrix between statements is complete (Figure 1c), the similarity between documents may be computed.  The document similarity measure selects the pairing of statements which maximizes the similarity between documents.  We also assume that pairings must be consistent (ie. a statement may be paired with only a single statement).  We define $\mathcal{D}$ as the set of all consistent pairings.  As the number of statements in each document may not be equal, some statements may not participate in a pairing.  Thus the similarity $S_{ik}$ between text entries $T_i$ and $T_k$ is defined as the following 

\begin{equation}
S_{ik} =  \max  ~ \left \{ \frac{ \sum_{(j,l) \in d} S_{ijkl} }  { \max \{ \mid T_i \mid, \mid T_k \mid \} }  \mid d \in \mathcal{D}_{ik} \right \}
\end{equation}

For example in Figure 1, the set of pairings would be ($T_{11}$,$T_{21}$) and ($T_{12}$,$T_{22}$), resulting a similarity of 0.55.  The similarity matrix is completed by repeating the process for each pairing of documents.

\section*{Supervised Textual Embedding}
In this section we present a supervised Laplacian eigenmap (SLE) method to enhance predictive models by embedding textual predictor into a low dimensional euclidean space which preserves the local similarities among textual data.  This method is presented in the following subsections: 1) we present Laplacian eigenmaps as an independent concept for dimensionality reduction for textual data, 2) we introduce a new supervised method and algorithm to augment base learners by incorporating Laplacian eigenmaps of textual predictors, and 3) we present methods to estimate the low dimensional representation of new data.

\subsection*{Laplacian Eigenmaps}
Laplacian eigenmaps are a geometrically motivated solution to developing low-dimensional euclidean representations of high-dimensional data such as text.  Laplacian eigenmaps preserve the local information optimality by causing similar points in the high dimensional space to be close in the low-dimensional representation.  We denote the low dimensional representation as $X = [X_1, X_2, \ldots, X_m]$, which consists of m points in $\Re^n$.   Laplacian eigenmaps assume there exists a similarity measure $\chi$ for the high dimensional data.  An effective similarity measure is essential to preserving the local structure within the embedding.  We assume the similarity matrix $S$ for the textual data is computed by our transformation process, where $S_{ij}$ denotes the similarity between documents $i$ and $j$.  Laplacian eigenmap's objective function, which is to preserve the local similarities within the low dimensional representation, is defined as follows 


\begin{equation}
	\Phi(X,S) = \sum_{i,j} \| X_i - X_j \|S_{i,j} = \mbox{tr}(X^TLX)
\end{equation}

where $L = D - S$, $D_{ii} = \sum_j S_{ij}$, and $L$ is symmetric positive semidefinite matrix that represents a Laplacian matrix.  The optimization function is 

\begin{eqnarray*}
X & = & \arg \min_{X} \{  \Phi(X,S) \mid X^TDX = I \} \\
\end{eqnarray*}

The constraint to the objective function prevents the solution from collapsing into a space less than $m-1$.  Without it, the optimal solution reduces to the trivial solution.  Standard methods show the solution is the matrix of eigenvectors corresponding to the lowest eigenvalues of the generalized eigenvalue problem $Lx = \lambda Dx$.  Alternatively, a solution may be found using methods such as gradient descent.  These methods may require the partial derivative of $\Phi(X,S)$.

\begin{equation}
 \frac{\partial tr(X^TLX)}{\partial X} = 2LX
\end{equation}

\begin{figure}
\centering
\setlength\fboxsep{0pt}
\setlength\fboxrule{0.8pt}

\fbox{\includegraphics[width=7.5cm]{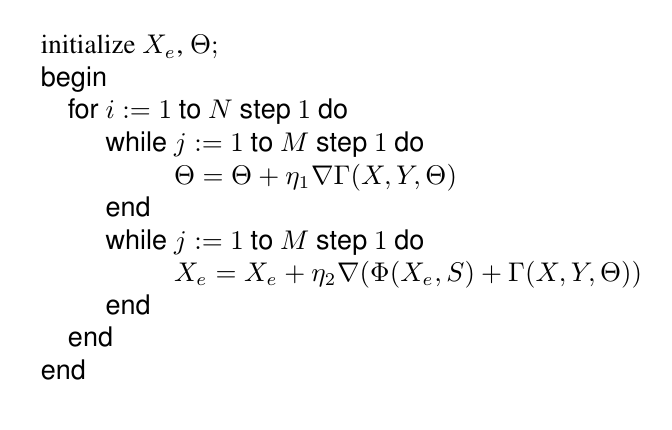}}
\caption{Supervised Laplacian eigenmap alternating optimization}
\end{figure}

\subsubsection*{Supervised Laplacian Eigenmaps}
In this section we introduce a new supervised method to jointly optimize the low-dimensional embedding and learn the predictive model's parameters. Unsupervised dimensionality reduction techniques, such as Laplacian eigenmaps, may construct low dimensional representations with low or no predictive ability.  Our supervised approach jointly optimizes the textual embedding and minimizes the base learner's loss to construct a low-dimensional representation of textual predictors with high predictive ability for a specific task.  Supervised Laplacian eigenmaps are defined as follows
  
\begin{equation}
\{ X_e, \Theta \} = \arg \min _{X_e \Theta} \Phi(X_e, S) + \lambda \mbox{Loss}(\Gamma(X, Y, \Theta))
\label{eq:jointopt}
\end{equation}

where $X \in \Re^{m \times n}$ and $X_e \in \Re^{m \times l}, X_e \in X$ represents the low dimensional embedding of text documents.  $S \in \Re^{m \times m}$ is the document similarity matrix and $Y \in \Re^{m \times 1}$ is the response variable.  We denote the base learner as $\Gamma(X, Y, \Theta)$,  where $\Theta$ is the set of learned parameters.  Equation \ref{eq:jointopt} may be solved by various methods depending on the selection of $\Gamma$. However we propose solving Equation \ref{eq:jointopt}, when the $\Gamma$ is differentiable in both $X_e$ and $\Theta$, using gradient descent to perform alternating optimization.

Our alternating optimization algorithm using gradient descent is defined in Figure 3.  This implementation with an appropriate $\lambda$, searches for a low-dimensional representation with potentially higher generalized predictive ability than unsupervised Laplacian eigenmaps.  Gradient descent is a first order optimization method which attempts to find a local minimum by moving in the direction of the negative gradient. By following the gradient with respect to $\Theta$, $\Theta$ converges to a critical point which is expected to be a local minimum.  This step optimizes the base-learner with respect to the observed data and current embedding.  The next optimization  is updating $X_e$ with respect to $\Theta$.  This optimization step updates $X_e$ while jointly preserving local textual similarities and embedding the response variable. The alternating optimization is repeated until the solution converges or reaches an iteration limit.  By performing alternating optimization, either $\Theta$ or $X_e$ potentially move out of their respective local minimum when optimizing with respect to the other variable.  As each optimization may improve the objective value and move the local minimum of the other variable, this allows $\Theta$ and $X_e$ to potentially converge to a more optimal value.  While this does not guarantee a global minimum, this method typically finds a more optimal solution than optimizing the variables independently.

One limitation of this method is the objective function is very sensitive to the value of $\lambda$;  an improper $\lambda$ may lead to an embedding which does not generalize. If the $\lambda$ value selected is "too small", the embedding will fully capture the response variable and the method will have poor generalized performance.  However, if the $\lambda$ is "too large", the embedding reduces to the trivial solution.  Determining the optimal $\lambda$ value may be difficult.  However we found success in initializing the embedding to the result of the unsupervised Laplacian eigenmap and performing optimization for a limited number of iterations with a $\lambda$ value considered "too small".

\subsection*{Estimating New Data Points}
After the predictive model has been trained, the low dimensional representation of new textual data must be estimated as our method is a supervised approach.  To estimate the embedding of new  data, we assume that the true embedding of a text entry can be proxied by the average low dimensional representation of its $k$ nearest neighbors (KNN) (Equation \ref{eq:average}), where the nearest neighbors are the most similar data points as defined by the similarity matrix $S$.  We define the set of nearest neighbors as $\mathcal{N}$.  We also explore an alternative weighted average (Equation \ref{eq:weighted}).  This method weights each neighbor by its similarity.

\begin{equation}
X_{e_i} = \frac{1}{p} \sum_{X_{e_j} \in \mathcal{N}_i } X_{e_j} 
\label{eq:average}
\end{equation}

\begin{equation}
X_{e_i} = \left\{ \begin{array}{rl}
 \frac{1}{\rho }\sum_{S_{ij} \in \mathcal{N}_i} S_{ij}X_{e_j} &\mbox{ if $\rho > 0$} \\
  0 &\mbox{ otherwise}
       \end{array} \right.
\label{eq:weighted}
\end{equation}
where $\rho  = \sum_{S_{ij} \in \mathcal{N}_i} S_{ij}$.


%% file: results.tex
\subsection*{Dataset}

We evaluated our algorithm on data from a large single-center pediatric cardiology practice.  The dataset contains anonymized records of 2,257 pediatric patients who underwent a clinical pathway focused on chest pain. This pathway consists of the following three components: a) patient and family history questionnaires completed prior to seeing physician and again reviewed with physician, b) comprehensive physical examination, and c) 12-lead electrocardiogram (EKG).

\begin{table}
\centering
\begin{tabular}{  l  l }
  \hline   
  {\bf Statistic} & {\bf }  \\   \toprule
  Total patients  & 2257  \\
  Patient age & 12.7 $\pm$ 3.9 \\
  Patients w/ cardiac disease & 25.1\%  \\   
  Patients w/ major cardiac disease  & 3.9\%  \\    
  Patients w/ positive physical exam & 37.2\% \\ 
  Patients w/ positive EKG & 20.7\%  \\ \hline
\end{tabular}
\caption{Basic patient statistics }
\label{basic-stats}
\end{table}

The patient and family questionnaires comprise the majority of the patient record.  These questions require the patient to accurately describe his/her condition and family history. Each question is in a yes/no or multiple choice format, with the exception of 2 questions: a text entry for the  patient's chief complaint (typically 2 - 10 words and contains 1 - 3 statements) and an additional question asking patients to diagram the location of chest pain. The latter question was excluded from the evaluation. The questionnaires were provided to patients in paper format and were later transcribed as EHRs.  We assume the errors introduced during transcription are minimal as free response question were limited. Patients who had not completed the questionnaires were removed from the data set, leaving 2,067 records remaining. The physical exam outcome was recorded as a free text entry written by the physician.  Each diagnostic test includes its CPT code and a text entry describing the outcome.  Diagnoses were encoded as both an ICD-9 code and a text entry.

In addition to the patient records, we received the diagnosis severity (no, minor, or major cardiac disease). The diagnosis was determined by a panel composed of physicians from this practice.  Each physician independently reviewed the diagnoses and assigned them one of the three possible outcomes.  If the reviewers' selections were unanimous, the action was considered valid.  However if the reviewers were not unanimous in their selection, discussion followed and a selection was made as a panel.  The panel identified 20 diagnoses as major cardiac disease,  14 as minor, and the remainder as no cardiac disease.  Examples may be seen in Table \ref{severity-sample}.

\begin{table}
\centering
\begin{tabular}{   l l l l    }
  \hline   
  {\bf ICD-9 } & {\bf Minor Cardiac Disease} & {\bf ICD-9 } &{\bf Major Cardiac Disease} \\  \toprule   
  427.61 & Supraventricular premature beats  &  426.7 &  Anomalous atrioventricular excitation  \\ 
  429.5 &  Rupture of chordae tendineae  & 745.54 & Anomalies of cardiac septal closure \\
  756.83 & Ehlers-Danlos syndrome & 746.02 & Stenosis of Pulmonary Valve, congenital \\
  785.1 & Palpitations &   746.85 & Coronary artery anomaly \\ 
  785.0 & Tachycardia, unspecified & 747.31 & Pulmonary artery coarctation and atresia \\ \hline
\end{tabular}
\caption{Examples of diagnosis severity}
\label{severity-sample}
\end{table}

\subsection*{Analysis}
We applied the proposed method, SLE, to the pediatric cardiology data to predict if patients were diagnosed with cardiac disease (major or minor).  To evaluate whether our supervised method for embedding textual predictor improved patient classification, we compared its performance with other unsupervised and supervised dimensionality techniques which were combined with a base classifier.  We implemented logistic regression with L2 regularization as the base classifier for all models.  The evaluation dataset included all 69 numerical features and a single text field, the patient's chief complaint.  We excluded the text results of the physical exam and EKG as many of the entries explicitly made mention of the outcome.

We used five-fold cross validation to assess the performance of all methods.  The results are presented as the average outcome from cross validation.  The discrimination of the models was evaluated by several metrics: the area under the receiver operating characteristic curve (AUC), Matthews Correlation Coefficient (MCC), sensitivity, and specificity. AUC assess a classifier's ability to balance the true positive and false positive rates.  New research indicates the AUC is a noisy classification measure\cite{hanczar}. We included MCC as it considers both true and false positive and negative rates. It is considered a more balanced measure than AUC, especially when classes are imbalanced. A MCC of 1.00 indicates perfect discrimination while a MCC 0.00 indicates no better than random classification.  Sensitivity and Specificity values were selected where the MCC achieved its highest value.

The unsupervised methods we implemented for comparison are unsupervised Laplacian eigenmaps, LSI, and LDA. We also compare the performance with the supervised dimensionality reduction technique LFDA. These methods utilized a term-document matrix generated from the chief complaint field.  Performance was also compared when using a TD-IDF term-document matrix with these methods.  We gave the unsupervised methods an unfair advantage in computing the the low dimensional representation.  These methods compute the embedding of all the textual data, including both test and training data, before training the base classifier.  This means the unsupervised methods do not suffer the performance degradation associated with estimating the embedding of test data.  Each classifier was initialized with a random seed and trained using the numerical and low dimensional embedding.  If any model during cross validation achieved a training AUC less than 0.65, the classifier was considered to have achieved a local optimum and the classifier was retrained with a new seed.  The low dimensional representation produced using Gibb's sampling LDA was computed using MALLET with 2,000 sampling iterations, $\alpha =0.9$, and $\beta = 0.01$.  We used Matlab's commands \textit{svd} and \textit{eig} to compute the SVD and Laplacian eigenmap respectively.  For LFDA, we used Sugiyama's implementation\cite{sug06,sug07} of kernel LFDA with $\sigma = 1.04$.  For each model, the optimal parameters were selected to maximize their respective MCCs. We estimated the embedding of test data for all supervised methods using the KNN methods which were proposed in the methods section.  The number of number neighbors was determined using cross validation.  The similarity matrix was computed using th transformation process which we presented.  The transformation vector $\omega$ was defined as follows: all transformations were defined to have a similarity value of one except the transform Missing which had a similarity of zero.  The transformation process utilized WordNet and some manual definitions for its dictionary.

\subsection*{Results}

\begin{figure}
\centering
\includegraphics[width=16.5cm]{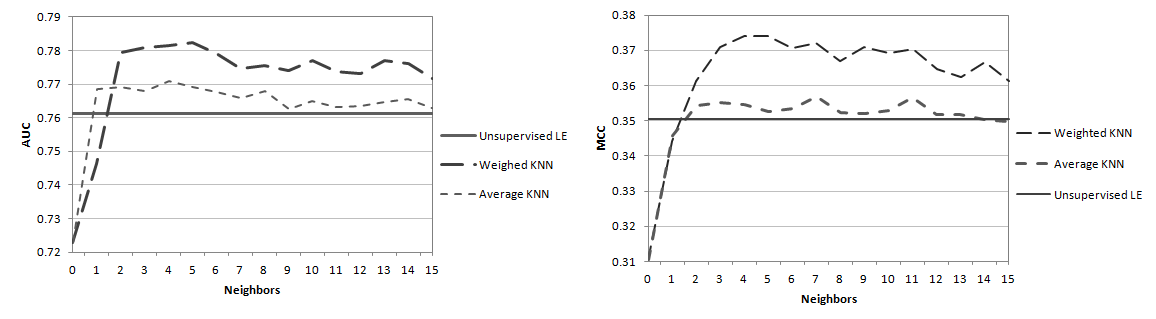}
\caption{Supervised Laplacian eigenmaps with $l = 20$ }
\end{figure}

Our evaluation includes two baselines: one excluding textual features to illustrate the potential of incorporating textual predictors, and another including them to illustrate the performance of SLE in comparison with alternative methods.  For the first baseline we applied logistic regression with L2 regularization to the numeric features of the pediatric cardiology dataset.  Five-fold cross-validation resulted in an AUC $= 0.723$ and MCC $= 0.310$.  For the second baseline we included the textual features by implementing unsupervised Laplacian eigenmaps and varying the number of dimensions in the low dimensional representation from zero to fifty. We found unsupervised Laplacian eigenmaps achieved its highest performance for both AUC and MCC at $l = 20$ dimensions, with AUC $= 0.761$ and MCC $ = 0.355$.  This is equivalent to a 5.26\% increase in AUC and a 14.5\% increase in MCC over the baseline which excludes textual data.  

For SLE, we determined which of our proposed methods is most effective for estimating new/test data.  We evaluated SLE using both average KNN (Equation 4) and weighted KNN (Equation 5) while varying the number of neighbors. We maintained the of dimensions, $l = 20$, at the optimal level from unsupervised Laplacian eigenmaps.  The AUC and MCC results are shown in Figure 4.  The results indicate that the both average and weighted average KNNs outperforms unsupervised Laplacian eigenmaps. Weighted KNN is consistently more effective then average KNN and achieves its highest performance at five neighbors. Figure 4 indicates weighted KNN's highest performance was AUC $= 0.782$ and MCC $= 0.374$ while average KNN achieved AUC $= 0.771$ and MCC  $= 0.357$.  The remainder of the evaluations for SLE used weighted five-nearest neighbors.

Having determined the optimal number of dimensions and neighbors for SLE, we compared the performance with the alternative unsupervised and supervised methods.  Each method was tuned for optimal performance and varied in the number of dimensions $l$.  The results of the comparison is shown in Figure 5.  As illustrated in the graphs, the maximal performance is dependent on the number of dimensions and metric.  As performance was similar yet improved, methods utilizing the term-document matrix used the TF-IDF weighting.  SLE and LSI collectively capture the highest performance for both metrics for almost the entire range of $l$.  Note that the unsupervised methods have an advantage by not estimating the embedding of textual predictors for test data.  LSI outperforms SLE for very low dimensions ($\le 17$ for AUC and $\le 15$ for MCC), however SLE outperforms LSI for higher dimensions.  When LSI estimates the embedding of test data using the same manner presented in this paper, LSI achieved AUC $= 0.751$ and MCC $= 0.349$, a loss of 0.031 and 0.011 respectively. LDA is consistently the lowest performing method while LFDA is nearly consistently the third highest performing method. LFDA is the highest performing on both metrics for extremely low dimensions ($\le 3$).  The highest performance for each method by sensitivity, specificity and likelihood ratios are included in Table 3.  SLE showed an increase in 8.16\% in AUC and 20.6\% in MCC over the baseline excluding the textual data and a 2.69\% and 5.35\% increase in AUC and MCC respectively over unsupervised Laplacian eigenmaps.

\begin{figure}
\centering
\includegraphics[width=16.5cm]{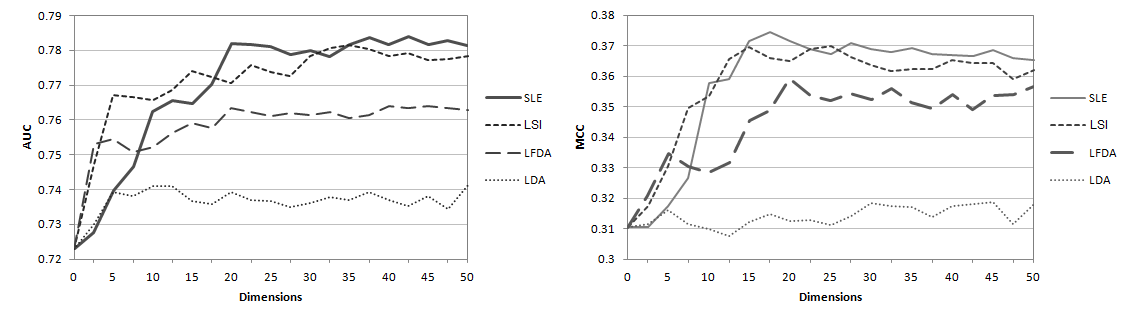} 
\caption{Comparison of dimensional reduction techniques for textual predictors}
\end{figure}


%% file: discussion.tex
\section*{Discussion}
Developing new techniques to leverage textual data in diagnostic aids is important to improving patient care.  Physician and patient notes often contain critical predictive information, such as pertinent patient history or chief complaints.  As illustrated by our results, augmenting diagnostic aids with textual data can substantially increase performance.  These improvements have a direct impact on improving resource utilization and patient outcomes.  Our unique method simultaneously allows a base learner to learn its parameters and discover the supervised low-dimensional representation of textual data.  

Another key advantage of our method is it abstracts away from a specific base-learner.  The method requires the base learner to differentiable in $\Theta$ and $X_e$ but still provides a large amount of freedom to select a base-learner which is most advantageous for the dataset.  This approach has many advantages over a method which fully integrates the textual dimensionality reduction with a specific classifier. 

Our evaluation indicates supervised Laplacian eigenmaps are an effective method for diagnostic aids within pediatric cardiology.  Our method outperformed the alternative methods in terms of AUC, MCC, and likelihood ratio positive.  LSI was clearly the most effective alternative and achieved similar, yet lower, performance to SLE.  It is interesting to note that LSI outperformed SLE for very low dimensions.  LSI likely achieves higher performance at low dimensions because the projections which maximize the sample variance also are correlated with high predictive information in our dataset. While LSI is able to outperform SLE for low dimensions, supervised Laplacian eigenmaps ultimately produced a more optimal embedding by incorporating the response variable.

A key area which differentiates supervised Lapacian eigenmaps from the SVD approach is the trade-offs the LSI faces between performance and speed. LSI obtains nearly comparable performance to our method using an unfair advantage of explicitly calculating the embedding of the test data points. This advantage is also a limitation which forces the method to recompute or update\cite{zhaUpdate} the SVD and retrain the learner for new data.  As an alternative, the LSI approach could reduce its complexity by sacrificing performance and estimate the embedding of new data points. When LSI uses the method presented in this paper to estimate data, SLE easily outperforms LSI as it incurs a loss of 0.031 AUC.

When comparing methods to estimate the embedding of new textual data, it is not surprising to find the weighted KNN method outperforms a traditional averaging approach.  Intuitively, the weighted averaging approach emphasizes the influence of the most similar neighbors.  Whereas each neighbor has equal influence in average KNN regardless of whether a nearest neighbor is consider "very" similar or only "somewhat" similar.  The weighted KNN method will estimate the embedding to be closer to the most similar points.


Our method does exhibit some limitations. SLE requires an accurate similarity matrix to construct an appropriate low dimensional representation.  Developing an accurate similarity measure for textual data may be difficult for applications which require significant domain knowledge. However, the method which we present is effective for concise clinical documents.  Alternative methods may be employed but such measures must consider complex textual relationships such as abbreviations, acronyms, and synonyms.  

An additional limitation arises from the implementation of our method.  The proposed objective function is very sensitive the value of $\lambda$.  As a solution we proposed an alternative optimization method.  While this may not be optimal, it was sufficient in practice.  Alternatively, $\lambda$ may be found through an incremental process which selects the $\lambda$ which leads to the highest generalized performance.  Despite these limitations, the potential improvement by incorporating low dimensional representations of textual data in diagnostic aids is evident.  SLE is an effective manner to incorporate textual data with a base predictor.

\begin{table}
\centering
\begin{tabular}{ l l l l l }
  \hline   
  {\bf Method } & {\bf Spec.} & {\bf Sens. } &{\bf Like. Ratio +} &{\bf Like. Ratio -} \\  \toprule   
  LDA & 0.390 & 0.858 & 2.744 & 0.711 \\
  SVD & 0.483 & 0.844 & 3.094 & 0.613 \\
  LFDA& 0.526 & 0.822 & 2.963 & 0.576 \\
  SLE & 0.470 & 0.851 & 3.149 & 0.623 \\ \hline
\end{tabular}
\caption{Classification performance}
\label{severity-sample}
\end{table}

\section*{Conclusion}
The unstructured nature of textual data presents a unique challenge for diagnostic aids to effectively integrate.  While many traditional machine learning techniques fail to incorporate text, this data may contain rich predictive information.  As a solution,  we presented a supervised Laplacian eigenmap method to embed textual predictors into a low dimensional euclidean space.  Improving machine-learning based diagnostic aids is a key element to improving patient outcomes. 

Our primary contribution is a supervised method which simultaneously utilizes dimensionality reduction to integrate text with traditional vector space data and learn a predictor’s parameters.  The advantage of supervised dimensionality reduction approach is in constructing a low dimensional embedding designed for a specific task.  Our evaluations, using a pediatric cardiology dataset, indicate our method outperforms supervised and unsupervised methods such as unsupervised Laplacian eigenmaps, SVD, LDA, and LFDA for both AUC and MCC.  However, LSI was the most competitive alternative to SLE, even outperforming for very low dimensional representations ($< \sim17$).  However, when LSI no longer has an unfair advantage and estimates the embedding of test data points, SLE easily outperforms LSI.  Our method extends to other unstructured or high dimensional data which can be effectively represented by a similarity matrix.  Our evaluation clearly demonstrates the potential and importance of incorporating textual data into diagnostic aids.  Further research on this topic is essential to reforming healthcare and improving outcomes.